\documentclass[letterpaper, 10 pt, conference]{ieeeconf}  

\IEEEoverridecommandlockouts                              

\usepackage{graphics} 
\usepackage{epsfig} 
\usepackage{mathptmx} 
\usepackage{times} 
\usepackage{amsmath} 
\usepackage{amssymb}  
\usepackage{multirow}
\usepackage{colortbl}  
\usepackage{booktabs} 
\usepackage{marvosym}

\usepackage{hyperref}  

\newcommand{\eg}{e.g.}
\newcommand{\ie}{i.e.}

\definecolor{shenghao_yellow}{RGB}{255,239,213}
\title{\LARGE \bf
Task-Oriented 6-DoF Grasp Pose Detection in Clutters}

\author{An-Lan Wang$^{1,\dagger}$, Nuo Chen$^{1,\dagger}$, Kun-Yu Lin$^{1}$, Yuan-Ming Li$^{1}$ and Wei-Shi Zheng\textsuperscript{1,2,\Letter}
\thanks{$^{1}$School of Computer Science and Engineering, Sun Yat-sen University, Guangzhou, China. 
Email: {\tt\small \{wanganlan, chenn65, linky5,liym266\}@mail2.sysu.edu.cn}}%
\thanks{$^{2}$Key Laboratory of Machine Intelligence and Advanced Computing, Ministry of Education, China.}%
\thanks{$^\dagger$ Equal Contribution}
\thanks{\textsuperscript{\Letter}Corresponding author: {\tt\small zhwshi@mail.sysu.edu.cn}}
}

\begin{document}

\maketitle
\thispagestyle{empty}
\pagestyle{empty}

\begin{abstract}
In general, humans would grasp an object differently for different tasks, \eg, ``grasping the handle of a knife to cut'' vs. ``grasping the blade to hand over''. 
In the field of robotic grasp pose detection research, some existing works consider this task-oriented grasping and made some progress, but they are generally constrained by low-DoF gripper type or non-cluttered setting, which is not applicable for human assistance in real life. 
With an aim to get more general and practical grasp models, in this paper, we investigate the problem named \textbf{T}ask-\textbf{O}riented \textbf{6}-\textbf{D}oF \textbf{G}rasp Pose Detection in \textbf{C}lutters (TO6DGC), which extends the task-oriented problem to a more general 6-DOF Grasp Pose Detection in Cluttered (multi-object) scenario. 
To this end, we construct a large-scale 6-DoF task-oriented grasping dataset, 6-DoF Task Grasp (6DTG), which features 4391 cluttered scenes with over 2 million 6-DoF grasp poses. 
Each grasp is annotated with a specific task, involving 6 tasks and 198 objects in total. 
Moreover, we propose One-Stage TaskGrasp (OSTG), a strong baseline to address the TO6DGC problem. 
Our OSTG adopts a task-oriented point selection strategy to detect \textit{where to grasp}, and a task-oriented grasp generation module to decide \textit{how to grasp} given a specific task. 
To evaluate the effectiveness of OSTG, extensive experiments are conducted on 6DTG. 
The results show that our method outperforms various baselines on multiple metrics. Real robot experiments also verify that our OSTG has a better perception of the task-oriented grasp points and 6-DoF grasp poses.
\end{abstract}

\section{INTRODUCTION}
Given the single-view point cloud of a cluttered scene, the general grasp pose detection problem requires a model to detect several grasp poses to grasp individual objects stably, which has achieved tremendous progress in recent years. 
Today's state-of-the-art grasping algorithms have shown great reliability and generalizability in grasping objects with high degree-of-freedom (DoF) gripper \cite{fang2020graspnet1billion, GSNet, sundermeyer2021contact, wu2025economic, wang2024single, lim2024equigraspflow}, under complicated environments. 
Several recent works start on solving grasp pose detection in some more challenging grasping scenarios, 
\eg,~ reactive grasping \cite{ marturi2019dynamicgrasping, akinola2021dynamicgrasping, yang2021reactivegrasping, liu2023reactivegrasping} 
and target-referenced grasping \cite{kurenkov2020targetoriented, liu2022targetoriented, li2022targetoriented}.

Despite these successes, there is still a significant gap between how robots pick and how humans grasp. 
The above-mentioned methods treat grasp as the final goal, focusing on how to grasp an object stably and reliably in different scenarios. 
However, when humans grasp an object, we generally want the grasp can help us finish a particular task. 
In other words, the grasp action is not the final goal, but a starting line towards further operations.

\begin{figure}[t]
    \centering
    \includegraphics[width=0.4\textwidth]{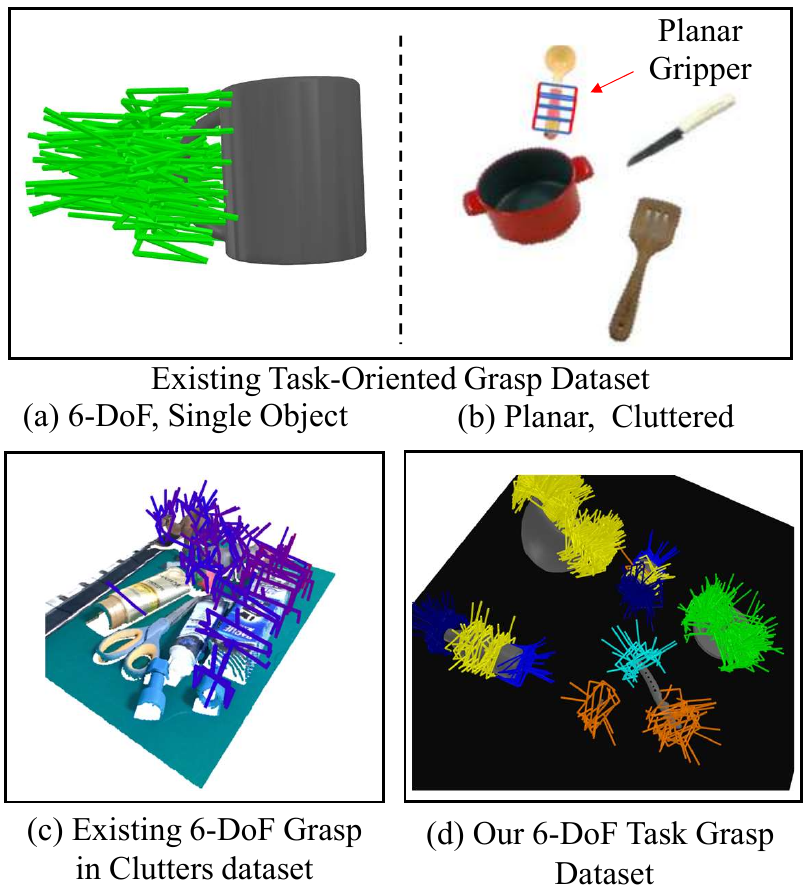}
    \caption{
    Examples of several existing datasets and our 6-DoF Task Grasp (6DTG) dataset. 
    Compared with the existing task-oriented grasping datasets (a) and (b), 6DTG uses 6-DoF grasp poses and cluttered scenes.
    Compared with the existing 6-DoF Grasping in clutters dataset (c), 6DTG provides the task annotation for each grasp.  
    Different grasp pose color represents different task in our 6DTG dataset.
    Best viewed in color.
    }
    \label{fig:compare_dataset}
\end{figure}

Recent works \cite{fang2020learningtog, murali2021same, sun2021gater, chen2022TOG, tang2023graspclip, tang2023graspgpt} provide a way to narrow this gap by investigating the task-oriented grasp pose detection problem, where a robot is required to detect a grasp pose to pick up an object to finish a particular task. 
However, these works are generally constrained by non-cluttered setting \cite{murali2021same,tang2023graspgpt} or low-DoF gripper \cite{fang2020learningtog, tang2023graspclip}, as shown in Figure \ref{fig:compare_dataset} (a) and \ref{fig:compare_dataset} (b). 
To perform household work in real life, an assistive robot must be able to perform \textbf{6-DoF} task-oriented grasping \textbf{in clutters}. 
Such as grasping a mug to fetch some water from a dinner table with a bowl, knife, and mug on top of it. 

To bridge this gap, we investigate the problem named \textbf{T}ask-\textbf{O}riented \textbf{6}-\textbf{D}oF \textbf{G}rasp pose detection in \textbf{C}lutters (\textbf{TO6DGC}). 
In this problem, a robot needs to detect several 6-DoF grasp poses to grasp an object (located in a cluttered scene) to finish specific tasks.
This problem is more realistic and practical compared with previous task-oriented grasp pose detection problems, which is a step closer to how humans grasp. 

TO6DGC problem has a higher requirement on datasets, \ie,~(1) each object should be able to perform different tasks and (2) each stable grasp needs to be annotated with its task label.  
Previous datasets, \eg, Graspnet-1Billion \cite{fang2020graspnet1billion} as shown in Figure~\ref{fig:compare_dataset}(c), is difficult to satisfy these requirements.
Therefore, by utilizing existing 3D objects and grasp annotation datasets \cite{savva2015shapenet, eppner2021acronym}, we construct a new large-scale dataset, named 6-DoF Task Grasp (6DTG) (as shown in Figure \ref{fig:compare_dataset}(d)). 
6DTG has several characteristics: (1) Fine-gained grasp task annotation. All grasps are annotated with their corresponding task to support task-oriented grasping. (2) 6-DoF grasp pose and cluttered scenes. This makes our dataset more practical compared with existing task-oriented datasets. (3) Densely annotated grasp poses. Dense grasp annotation has been proved to be important for training a robust grasp pose detection network \cite{fang2020graspnet1billion, eppner2021acronym, sundermeyer2021contact, zheng2025diffuvolume}.

\begin{figure}[t]
    \centering
    \vspace{0.2cm}
    \includegraphics[width=0.45\textwidth]{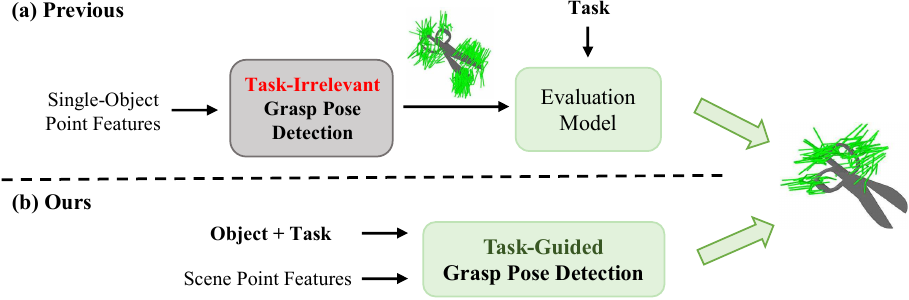}
    \caption{Previous methods \cite{chen2022TOG, tang2023graspgpt} focus on single object scenarios, and use a two-stage pipeline, \ie,~ generate task-irrelevant (stable) grasps firstly and then use an evaluation model to evaluate whether the grasp is suitable for a particular task. 
    In contrast, we propose a novel one-stage task-guided grasp pose detection model to detect task-oriented grasp in a holistic way.}
    \label{fig:grasp_pose_generation_comparison}
    \vspace{-0.2cm}
\end{figure}

Previous 6-DoF task-oriented grasping methods \cite{murali2021same, tang2023graspgpt}, generally focus on single object scenarios, adopting a two-stage pipeline. 
They first detect task-irrelevant (stable) grasps and then use an evaluation model to judge whether each grasp pose is suitable for the desired task, as shown in Figure \ref{fig:grasp_pose_generation_comparison}(a). 
Such a pipeline is time-consuming (detect and evaluate all stable grasps) and suffers from error accumulation, leading to poor evaluation precision. 
These methods perform even worse in solving the TO6DGC problem, as shown in the experiments. 
In this work, we further propose a strong baseline One-Stage TaskGrasp (OSTG) to address the TO6DGC problem, which \textit{\textbf{directly}} detects the task-oriented 6-Dof Grasp pose in a one-stage manner, as shown in Figure~\ref{fig:grasp_pose_generation_comparison}(b). 
Specifically, Our OSTG first adopts a task-oriented point selection module to localize the points that can be grasped to finish the desired task, namely \textit{where to grasp}, as we find that the task characteristics of a grasp pose is partially related to the grasp point. 
Based on the selected task-oriented points, we propose a novel task-guided grasp pose detection module that directly detects task-oriented 6-DoF grasp rotation, \ie, \textit{how to grasp}. 
We conduct extensive experiments on the 6DTG dataset, and the results demonstrate that our proposed OSTG model outperforms various baselines by a large margin.
Real robot experiments also verify the effectiveness of our OSTG model.

\section{Related Works}

\subsection{6-DoF Grasp Pose Detection in Clutters.}
6-DoF (or full-DoF) grasp pose detection in clutters, is a fundamental problem in robotics \cite{suarez2018can, dantam2016incremental, xu2024dexterous, gao2024riemann}, and is an important step toward Artificial General Intelligence \cite{shao2020manipulation, avigal2022manipulation, papallas2022manipulation, wang2023event, wang2024pargo, li2024egoexo, chen2024motiongrasp, li2024techcoach, lin2024diversifying, tang2022few, zhao2024harmonizing, tang2024mtvqa, tang2024textsquare, zhao2024tabpedia, lu2024bounding, shan2024mctbench, feng2024docpedia, du2024weakly}. 
Two different approaches are explored in this field. 
The first approach \cite{varley20156-DofGraspoing, GPD} adopts a sampling-evaluation strategy, which uniformly samples grasp candidates in the scene and then evaluates each grasp candidate using a deep network. 
The second approach \cite{6-dofview, fang2020graspnet1billion, ni2020pointnet++grasp, GSNet} adopts an end-to-end network. 
However, all above-mentioned methods focus on grasping objects stably and reliably, without considering the task associated with each detected grasp pose. In this work, we extend these to the task-oriented setting, which is more natural and practical.
\subsection{Task-Oriented Grasp Pose Detection}
For task-oriented grasp pose detection, previous research can be mainly divided into two categories: planar-based \cite{fang2020learningtog, sun2021gater, tang2023graspclip} and 6-DoF-based \cite{SG14000, murali2021same, chen2022TOG, tang2023graspgpt}. 
Research in the first category mainly took RGB (or RGB-D) images as inputs and outputs a set of \textbf{rotated bounding boxes} to represent the grasp poses. 
Due to the limitation of low DoF, their applications are restricted. 
Another line of research focuses on 6-DoF (full-DoF) grasp poses. 
However, they mainly focus on how to grasp single-object with specific tasks, which strictly restricts their application, \eg,~GraspGPT \cite{tang2023graspgpt}. 
In this work, we investigate the problem named task-oriented 6-DoF grasp pose detection in clutters, which is more realistic and practical than all previous ones, facilitating the development of task-oriented grasp pose detection. 
\subsection{Grasp Point Sampling Strategies}
To detect available 6-DoF grasps in cluttered scenes, direct regression in high-dimensional discontinuous grasp space is quite challenging \cite{mousavian20196dofgraspnet}. 
Several works propose to sample grasp points first and then detect grasp poses based on the sampled points. 
GPD \cite{GPD} and PointNetGPD \cite{pointnetGPD} use a simple uniform sampling strategy to select points. 
Contact-GraspNet \cite{sundermeyer2021contact} uses the contact point of the gripper and object to effectively represent the grasp.
GSNet \cite{GSNet} proposes the concept of graspable point sampling, achieving outstanding performance in cluttered scene 6-DoF grasping. 
In this work, we propose a task-oriented point selection strategy to handle the task-oriented 6-DoF grasp pose detection in clutters problem. 

\section{Task-Oriented 6-DoF Grasp Pose Detection in Clutters}
\subsection{Problem Definition}
In general 6-DoF grasp pose detection in clutters, given the single-view point cloud $P$ of a clutter, the model needs to detect several grasps $G$ to grasp individual objects stably. Each grasp $g \in G$ is represented by $(R, T) \in SE(3)$, where $R \in SO(3)$ and $T \in \mathbb{R}^3$ are the rotation and translation of grasp $g$. 
$SE$ is the Special Euclidean group, and $SO$ is the Special Orthogonal group.
For the task-oriented setting, we additionally provide an object class $o$ and a task class $t$, the model needs to detect several task-oriented grasps $G^*$, which can be formulated as a posterior probability :
$p\left ( G^*| P, o, t \right )$,
where $G^*$ contains all successful grasps $g^*$ that can grasp object $o$ to finish the task $t$. 
This problem is more challenging than previous task-oriented settings (\ie,~cluttered scenes \textit{vs.} single-object, 6-DoF grasp \textit{vs.} planar grasp ). 
\begin{table}[t]
    \centering
    \vspace{0.2cm}
    \caption{Object categories and corresponding tasks in our dataset}
    \resizebox{0.45\textwidth}{!}{
    \begin{tabular}{c|c|c|c|c|c|c}
    \hline
     & Mug & Bottle & Knife & Hat & Bowl & Scissor \\ 
    \hline
        \multirow{2}*{Tasks}  & Grasp, Wrap,   & Grasp, Wrap,   & Handover,  & Grasp,  & Grasp,  & Handover, \\
         &  Pour, Contain & Contain & Cut & Wear & Wrap & Cut\\
        \hline
    \end{tabular}}
    
    \label{tab:object_task_relation}
\end{table}

\begin{table}[t]
    \centering
    \vspace{-0.2cm}
    \caption{Comparison of our 6DTG with previous task-oriented grasping Datasets. Our dataset is the first to provide 6-DoF task-oriented grasp annotation in clutters.}
    \resizebox{0.5\textwidth}{!}{
    \begin{tabular}{c|cccccc}
    \hline
     \multirow{2}{*}{Datasets} & Object & \multirow{2}{*}{Objects} & \multirow{2}{*}{Tasks} & Grasps &  6-DoF & \multirow{2}{*}{Clutters} \\ 
     & Category & & & Number&Grasp&\\
    \hline
        SG14000 \cite{SG14000} &  5 & 44 & 7 & 12K & \checkmark  &  \\
        TaskGrasp \cite{murali2021same} &  75 & 191 & 56 & 250K & \checkmark  &  \\
        GATER \cite{sun2021gater} &  - & 83 & 10 & 12K &  &  \\
        GraspCLIP\cite{tang2023graspclip} &  28 & 96 & 38 & - &  & \checkmark \\
        AGD \cite{chen2022TOG} & 6 & 203 & 6  &  100K & \checkmark &  \\
        \rowcolor{shenghao_yellow} \textbf{6DTG(Ours)} & 6 & 198 & 7 & \textbf{2,000K} & \textbf{\checkmark} & \textbf{\checkmark} \\
        \hline
    \end{tabular}}
    \vspace{-0.3cm}
    \label{tab:dataset_comparison}
\end{table}
\subsection{6DTG Dataset}

\vspace{0.2cm}
\noindent\textbf{- Data Collection}

\textbf{(1) Objects and stable grasps prototype.} 
In this work, our focus lies specifically on kitchen and household objects that can be grasped to operate different tasks. 
Taking this into account, we choose ShapeNetSem~\cite{savva2015shapenet} and ACRONYM \cite{eppner2021acronym} as the objects and stable grasp prototype source, respectively.
After human selection, 
We choose 6 object categories and 198 objects, as shown in Table \ref{tab:object_task_relation}. 

\textbf{(2) Scene generation.} Following ACRONYM \cite{eppner2021acronym}, we generate cluttered scenes that multiple objects are placed on top of a support table using trimesh. 
Firstly, we randomly select 3-6 objects that cover different object categories. 
Secondly, given the support table, we sequentially sample locations from a 2D Gaussian (centered around the table's center) to place the selected objects on top of it, while ensuring the resulting configurations of all objects are collision-free. 
It's worth mentioning that, for small objects like knives and scissors, we additionally place a small cube under them to lift it for $5 cm$. 
This adjustment is made to facilitate grasping, as directly picking up scissors from the tabletop is challenging for a parallel-yaw gripper. 
For each generated scene, we render two point-cloud observations using different camera poses.
\begin{figure}[t]
    \vspace{0.2cm}
    \centering
    \includegraphics[width=0.95\linewidth]{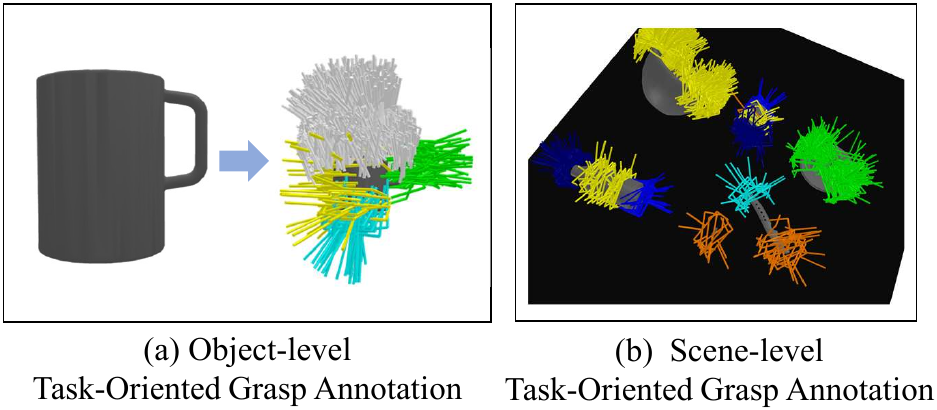}
    \caption{Visualizations of (a) Object-level and (b) scene-level task-oriented grasp annotation. Different colored grippers represent different tasks. Best viewed in color.}
    \vspace{-0.2cm}
    \label{fig:Grasp_Annotation}
\end{figure}

\begin{figure*}[t]
    \centering
    \vspace{0.2cm}
    \includegraphics[width=0.9\textwidth]{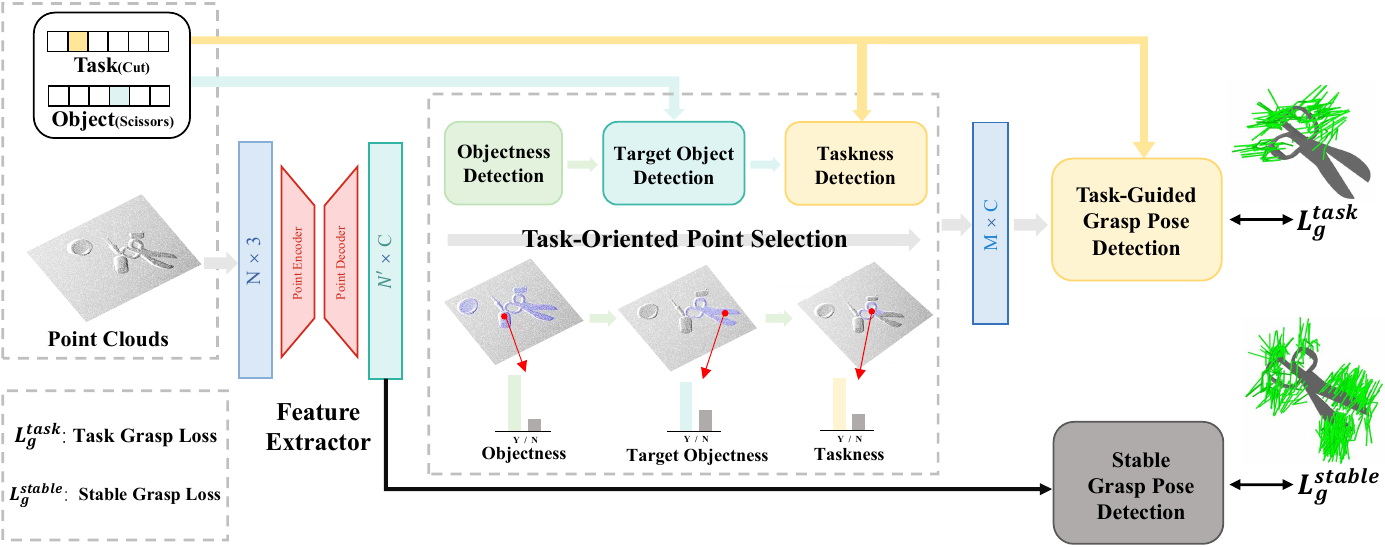}
    \caption{Overview of our proposed One-Stage TaskGrasp (OSTG) model. 
    In the figure, we provide an example of a model detecting grasp poses to grasp the \textit{scissors} in the clutters that can finish the \textit{cut} task. 
    First of all, the point encoder-decoder processes $N\times3$ points and outputs $N' \times C$-dim point features. 
    Supervised by the point-wise labels, the Task-oriented Point Selection module selects $M$ points according to objectness, target objectness, and taskness, in a step-by-step manner. 
    These task-related grasp point features are then fed into the Task-Guided Grasp Pose Detection module to generate task-oriented grasps. 
    We additionally add a stable grasp loss to supervise the model.
    Best viewed in color.
    }
    \label{fig:pipeline}
    \vspace{-0.3cm}
\end{figure*}

\vspace{0.2cm}
\noindent\textbf{- Annotations}

Next, we illustrate how to generate object-level and scene-level grasp task labels
, as shown in Figure \ref{fig:Grasp_Annotation}.

\textbf{(1) Object-level.}  Inspired by AGD \cite{chen2022TOG} and 3D AffordanceNet~\cite{deng2021affordancenet}, we first annotate all successful grasps of objects with different task labels. 
The task classes we use are the same as 3D AffordanceNet~\cite{deng2021affordancenet}, with details of the tasks each object can perform outlined in Table \ref{tab:object_task_relation}.
For each object, we first use a pre-trained 3D AffordanceNet \cite{deng2021affordancenet} to detect the object affordance part, and then each grasp is labeled according to the grasp point. 
Applying task labels according to the grasp point can help identify the task property of a grasp, but cannot serve as the final judgment. 
Therefore, we further manually verify whether the grasp for the object is suitable for the task. 
Finally, we got about 300K object-level task annotated grasps, an example is shown in Figure~\ref{fig:Grasp_Annotation}(a).

\textbf{(2) Scene-level.} As our dataset is collected from simulation, it's easy to use the object-level grasp task labels to label the scene-level grasp, using the object poses. 
It's worth noting that grasps colliding with any scene geometry is removed. An example is shown in Figure~\ref{fig:Grasp_Annotation}(b).

\vspace{0.1cm}
\noindent\textbf{- Data Triplet Generation}

After the above processes, we generate data triplets $(P, o, t)$ by scanning all rendered scenes, where $P, o, t$ is the point cloud, object class, and task class. 
Totally, we have 16042 triplets for training and 6841 triplets for test. 

\subsection{Comparison with Related Datasets}
In Table \ref{tab:dataset_comparison}, we compare our 6DTG with existing task-oriented datasets. 
Our dataset is the first task-oriented dataset that focuses on both \textit{cluttered scenes} and \textit{6-DoF grasp poses}. 
Additionally, we emphasize that our dataset provides a much larger amount of grasp annotation, compared with the previous datasets, whether planar-based or 6-DoF-based.

\section{Method}
Existing task-oriented grasping methods, generally focus on 1) planar-based grasp in clutters, and 2) 6-DoF-based grasp for a single object. These methods can not handle the challenging TO6DGC problem. 
\subsection{Grasp Representation}
In the TO6DGC problem, each successful grasp in the high dimensional SE(3) space further corresponds to a task.
So it is difficult for a learning-based model to generate the distribution of successful 6-DoF task-oriented grasps. 
Therefore, we introduce the grasp representation in our work, which can help solve this problem. 

Specifically, we decompose a 6-DoF grasp pose into (1) the corresponding task-oriented grasp point, (2) the 3-DoF grasp rotation, and (3) the grasp width of a parallel-yaw gripper. 
The grasp rotation $R_g$ is further decomposed into an approaching direction $a\in \mathbb{R}^3$, and a baseline direction $b\in \mathbb{R}^3$, as shown in Figure \ref{fig:gripper}. 

Therefore, we can first predict a suitable grasp point for the desired task and then estimate the corresponding 3-DoF rotation and grasp width, which greatly eases the TO6DGC. 

Based on this, we propose our \textbf{O}ne-\textbf{S}tage \textbf{T}ask \textbf{G}rasp (OSTG) model. 
Different from previous two-stage methods that first generate all stable grasps and then evaluate each grasp, our OSTG \textit{\textbf{directly}} generates task-oriented grasp poses \textit{\textbf{in one stage}}.

\subsection{Model Overview} Figure \ref{fig:pipeline} illustrates the proposed One-Stage TaskGrasp (OSTG) model. 
Given the point cloud $P$, task $t$, and object $o$, we first use a point set backbone network PointNet++ \cite{qi2017pointnet++} to process the point cloud, which tasks in the raw point cloud with size $N \times 3$, and outputs features in the shape of $N' \times C $. 
Subsequently, the Task-Oriented Point Selection module learns the point that can be grasped to finish the desired task, namely \textbf{\textit{where to grasp}}. 
Finally, these selected task-related grasp point features, together with the task class one-hot vector, are fed into a Task-guided Grasp Pose Generation (TGPG) module, which directly generates task-oriented grasp poses, \ie, \textbf{\textit{how to grasp}}. 
We additionally add a stable grasp detector to predict stable grasp poses for efficient training.

\subsection{Task-oriented Point Selection}
First of all, we introduce the Task-oriented Point Selection module of our approach, which learns the point that can be grasped to finish the desired task, namely \textit{where to grasp}. 
The Task-oriented Point Selection (TPS) module samples the task-related points in a step-by-step manner. 
In specific, the TPS module samples points according to objectness (\ie,~whether the point belongs to an object), target objectness (\ie,~whether the point belongs to the target object), and taskness (\ie,~whether the point in the target object is suitable for the desired task). 

Formally, given the point features outputted by the backbone $f_{ori}$, we add three binary classification heads $h_o$, $h_{to}$, $h_{task}$ to select points, which can be formulated as: 
\begin{equation}
    f_o = {h_o}(f_{ori}),\ \ f_{to} = h_{to}(f_{o}, o),\ \ f_{task} = h_{task}(f_{to}, t),
\end{equation}
where $f_{o}$, $f_{to}$, and $f_{task}$ are the selected point features according to objectness, target objectness, and taskness respectively, and $o$, $t$ is the target object and task.
This point selection operation is finished step-by-step, and points are \textit{sampled (or oversampled)} to a fixed size after each head to support parallel computing. 
Such a step-by-step strategy has several benefits: 
(1) The points filtered by objectness or target objectness are not taken into further consideration, 
(2) As the point number decreases, the computation also reduces.

\subsection{Task-guided Grasp Pose Detection}
In this section, we propose a Task-guided Grasp Pose Detection module to learn the grasp rotation based on the selected task-oriented point feature and the task information. 

Specifically, the TGPD module takes two inputs, \ie, the selected task-oriented point features $f_{task}$ and the target task $t$. 
To guide the grasp parameter prediction, we first concatenate the task embedding to the point feature, which is given as follows:
\begin{equation}
    \Tilde{f}_{task}=Concat(f_{task}, t).
\end{equation}
After getting the concatenated feature, we then use three heads to predict two grasp rotation vectors, \ie,~the approaching vector $\Tilde{a}$ and the baseline vector $\Tilde{b}$, and the corresponding grasp confidence. 
As our grasp representation defined, the two vectors $a$ and $b$ should be orthonormal, we use a Gram-Schmidt orthonormalization to process $\Tilde{a}$ and $\Tilde{b}$, and then compute the grasp rotation matrix. 

Only using task-oriented grasps as supervision, it is challenging to train the overall model as the ground-truth task-oriented grasp pose number is limited. 
Therefore, we additionally add a task-irrelevant (stable) grasp pose detection module. 
This module takes in the point features from the feature extractor, outputs all the grasp poses, and is supervised with all available grasps. 
\begin{figure}
    \centering
    \vspace{0.2cm}
    \includegraphics[width=0.2\textwidth]{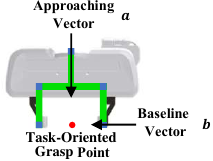}
    \caption{We represent a grasp rotation using an approaching vector and baseline vector. 
    To calculate the grasp loss, we further represent a grasp pose using \textcolor{blue}{five points} as shown above. 
    Best Viewed in color.}
    \label{fig:gripper}
    \vspace{-0.4cm}
\end{figure}

\subsection{Training Losses}
The overall training losses can be divided into two parts, \ie, the point selection loss $L_{point}$ and grasp loss $L_{grasp}$.

\vspace{0.1cm}
\textbf{Point Selection Loss.} As the task-oriented point selection module contains three steps, \ie, objectness, target objectness, and taskness. 
Each step is supervised with its ground-truth label and contributes a loss item:
\begin{equation}
    L_{point} = L_o + L_{to} + L_{task}.
\end{equation}

\vspace{0.1cm}
\textbf{Grasp Loss.} Following previous work \cite{sundermeyer2021contact}, we project a gripper grasp pose into 5 points, as shown in Figure \ref{fig:gripper}. 
Then, we calculate the loss of predicted grasps and their corresponding ground-truth grasp, which is defined as 
\begin{equation}
L_g = \frac{1}{n}\sum_{i}^{n} \widehat{s}_i\underset{u}{min}||g_i^{pred} - g_u^{gt}||,
\end{equation}
where the subtraction of two grasp poses is defined as the distance of the five corresponding points in 3D space. 
Therefore, the grasp loss is defined as:
\begin{equation}
    L_{grasp} = L_{g}^{task} + L_{g}^{stable},
\end{equation}
where $L_{g}^{task}$ is the task-oriented grasp loss and $L_{g}^{stable}$ is the stable grasp loss.

\textbf{Grasp Loss.} The training loss function is defined as:
\begin{equation}
    L_{overall} = L_{point} + L_{grasp}.
\end{equation}


\section{Experiments}

\subsection{Baselines}
Previous task-oriented grasping methods, generally focus on 1) planar-based grasp in clutters, and 2) 6-DoF-based grasp for a single object. 
It's challenging to adapt these methods directly to solve the TO6DGC problem (6-DoF and Clutters). We design two baselines. 

\vspace{0.1cm}
\textbf{Two-stage Baseline 1}. 
For this baseline, we adapt previous evaluation-based methods \cite{murali2021same, tang2023graspgpt}, which first generate all stable grasps and then evaluate each grasp.
In practice, for simplicity, we directly use the ground-truth stable grasp as the stable grasp, which can be seen as an upper bound.  
We use TaskGrasp~\cite{murali2021same} to evaluate whether each grasp is suitable for the task $t$ and object $o$. 
This baseline is constructed to verify whether previous single-object two-stage methods can handle the multi-object (cluttered) scenario. Additionally, we add a \textit{Random} baseline that randomly selects a task class from all task classes. 

\vspace{0.1cm}
\textbf{One-stage Baseline 2}. 
In this setup, we use the Contact-Graspnet \cite{sundermeyer2021contact} as the model to generate grasps. 
Different from its origin implementation, we simply concatenate a task one-hot and an object one-hot to the point features before it is fed to the final task-oriented grasp pose prediction head, which is supervised with the corresponding task-oriented grasp poses. 
We also keep the origin stable grasp prediction head, supervised with all stable grasp poses in the scene. 
We construct this baseline to verify whether a current state-of-the-art stable 6-DoF grasp pose detection model can handle the TO6DGC problem with a simple modification.

\subsection{Metrics}
For the one-stage methods, we combine previous 6-DoF grasping \cite{sundermeyer2021contact} and task-oriented \cite{chen2022TOG} works, and evaluate the performance using two metrics. 
(i) Coverage Rate. 
This metric only required the model to output the grasp poses that grasp the task-oriented points, without requiring the grasp pose to grasp the object stably.
(ii) Success Rate, which considers the grasp quality of each task-oriented-point grasp pose. 
This metric considers whether the grasp pose can grasp the object for the given task, Success Rate is set to 1 if at least one detected grasp pose satisfies the task-oriented requirement. 
To determine whether a predicted grasp pose $\Tilde{g}$ satisfies the task-oriented requirement, we calculate the grasp pose distance \cite{fang2020graspnet1billion} between $\Tilde{g}$ and all ground-truth task-oriented grasp poses. 
The grasp pose distance is defined as:
\begin{equation}
    D(G_1, G_2)=(d_t(G_1, G_2), d_\alpha(G_1, G_2)),
\end{equation}
where $G_1$ and $G_2$ are two grasp poses, $d_t(G_1, G_2) = ||t_1 - t_2||$ denotes the translation distance and $d_\alpha(G_1, G_2)=arccos\frac{1}{2}(trace(R_1, R_2^T)-1)$  is the rotation distance. 
$t_i$ and $R_i$ are the translation vector and rotation matrix. 
Based on this, one predicted grasp $\Tilde{g}$ is seen as positive if it has a $D < (th_d,th_\alpha)$ with at least one ground-truth task-oriented grasp. 
In this paper, we set $th_d = 3 $ cm and $th_\alpha = 30$ degree, as previous work \cite{fang2020graspnet1billion} done. 
For the two-stage methods, we compute the average grasp task classification precision as the Success Rate. 

\subsection{Comparison with Baselines}
In this section, we compare our OSTG model with several baselines on the proposed 6DTG dataset. 
As depicted in Table \ref{tab:main_experiment}, the previous two-stage evaluation-based method exhibits poor performance on our 6DTG datasets, even provided with the ground-truth stable grasp. 
This is because TaskGrasp \cite{murali2021same} is designed to evaluate grasp poses of single-object point clouds. 
But in our 6DTG dataset, the point cloud of cluttered scenes contains multi-object and background points, which is much more challenging than single-object. 
Additionally, it is worth noting that such evaluation-based methods take a long time to evaluate compared with our One-Stage TaskGrasp (OSTG), \ie, 8 hours \textit{vs.} 40 minutes. 

Compared with the one-stage Baseline 2, OSTG obtains significant improvement, \ie, $38.19\%$ in Coverage Rate and $5.32\%$ in Success Rate. 
The results demonstrate that our OSTG model effectively learns where and how to grasp, which is attributed to our proposed task-oriented point selection and task-guided grasp pose detection module. 

\begin{table}[t]
    \renewcommand\arraystretch{1.3}
    \centering
    \vspace{0.2cm}
    \caption{Comparisons with several constructed baselines on 6DTG dataset. $\dagger$ uses ground-truth stable grasp as input.}
    \resizebox{0.45\textwidth}{!}{
    \begin{tabular}{c|c|c|c}
        \hline
        & \multirow{2}{*}{Methods} & \multirow{2}{*}{Coverage}  & \multirow{2}{*}{Success Rate } \\
        &  &     &  \\
        \hline
        
        \multirow{2}{*}{Two-Stage$^\dagger$} 
        & Random     & - & 11.73    \\
        & Baseline 1 & - & 25.32    \\
        \hline
        \multirow{2}{*}{One-Stage} 
        & Baseline 2 & 5.39 & 66.81\\
        & \textbf{Ours}       & \textbf{43.58} & \textbf{72.13} \\ 
         \hline
    \end{tabular}}
    \vspace{-0.3cm}
    \label{tab:main_experiment}
\end{table}

\begin{table}[t]
    \centering
    \renewcommand\arraystretch{1.2}
    \caption{Ablation study of the main components of our method on 6DTG dataset. }
    \resizebox{0.5\textwidth}{!}{
    \begin{tabular}{c|cc|cc}
        \hline
        Model & TPS &  TGPD  & Coverage & Success Rate \\
        \hline
        Baseline &  &   &  0.2&  61.42\\
         + TPS & \checkmark & & \underline{40.19}   &  62.27 \\
         Ours & \checkmark  &  \checkmark   & \textbf{43.58}  & \textbf{72.13} \\
         
        \hline
        Ours w/o TPS &   & \checkmark    &  5.39 & \underline{66.81}\\
        
        \hline
    \end{tabular}}
    \label{tab:ablation}
    \vspace{-0.4cm}
\end{table}

\subsection{Ablation Study}
\textbf{Effect of main components in OSTG. } 
In Table \ref{tab:ablation}, we analyze the effect of each component of our proposed method. 
We directly use a state-of-the-art 6-DoF grasp pose detection model Contact-Graspnet as our baseline. 
By adopting a Task-oriented Point Selection module (TPS), we obtain a huge improvement in terms of coverage rate, \ie, $40.17\%$, which demonstrates the effectiveness of our proposed point selection strategy. 
Then, by introducing the Task-guided Grasp Pose Detection module (TGPD), our model obtains an improvement in Success Rate, \ie, $9.86\%$. 
In addition, if the TPS module is removed from the full model (the model degrades to the above-mentioned Baseline 2), the performance drops both in Coverage and Success Rate but still achieves better performance than the baseline. 

\begin{figure}
    \vspace{0.2cm}
    \centering
    \includegraphics[width=0.4\textwidth]{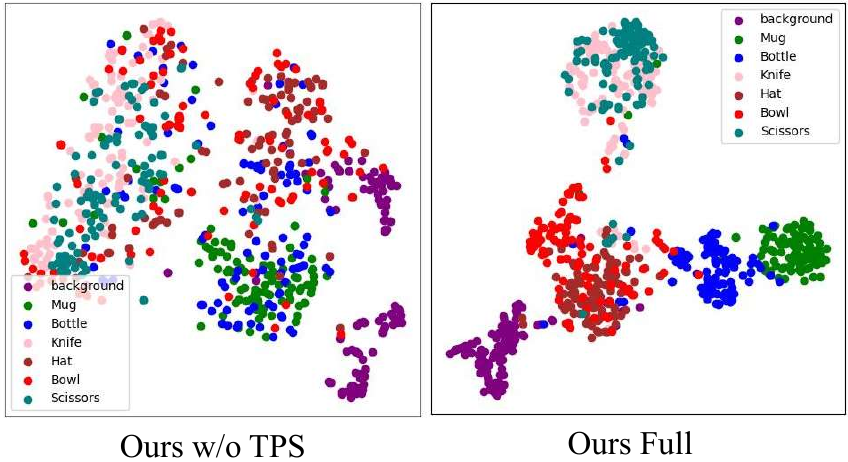}
    \caption{Analysis of the Task-oriented Point Selection (TPS) module using t-SNE \cite{van2008tsne}. 
    For each class, we randomly choose 100 point features. 
    }
    \label{fig:tSNE_visualization}
    \vspace{-0.4cm}
\end{figure}

\textbf{Qualitative analysis. }
Here, 
to verify the effectiveness of the proposed Task-oriented Point Selection (TPS) module, we employ t-SNE \cite{van2008tsne} to project the point features into 2D space. 
For every point class, we randomly choose 100 points for visualization, the results are shown in Figure \ref{fig:tSNE_visualization}. 
As we can see, ``Ours w/o TPS'' can hardly separate points that belong to different classes, resulting in all points being mixed within the 2D space.
In contrast, the point features extracted by our full model are grouped by the object labels, which suggests that our model has learned roughly distinct regions in the point feature space to correspond to each object point.

\begin{table}[t]
    \centering
    \renewcommand\arraystretch{1.4}
    \vspace{0.2cm}
    \caption{Real-world experiments.}
    \resizebox{0.45\textwidth}{!}{
    \begin{tabular}{c|c|c|c|c|c|c}
    \hline
     & \multicolumn{2}{c|}{Mug}  & \multicolumn{2}{c|}{Knife} & \multicolumn{2}{c}{Scissor} \\ 
    \hline
        Tasks  & Grasp  & Pour   & Grasp  & Cut  & Grasp  & Cut \\
    \hline
        Success/Total & 8/10 & 6/10 & 5/10 & 4/10 & 4/10  & 3/10 \\
        \hline
    \end{tabular}}
    \label{tab:realrobot}
    \vspace{-0.4cm}
\end{table}

\subsection{Real-World Experiment}
As the contributed dataset 6DTG is collected from simulation environments. 
We conduct real-world robot grasping experiments to verify whether our model trained from simulation dataset can transfer well to real-world grasping.

Specifically, we construct cluttered scenes by randomly placing some objects from knives, Scissors, Mugs, Bottles, and bowls, which are all unknown to the model. 
Then, we use the model to generate task-oriented grasps given a specific task and a target object. 
To perform the grasp, we use a fixed Franka Emika robot hand with a parallel-yaw gripper, and a third-view Kinect Azure RGB-D camera is placed on the right front of the robot. 
As shown in Table~\ref{tab:realrobot}, for easy tasks like "grasping an object", our model predicts execution grasp with a high success rate. 
For more challenging tasks, \eg, grasping the knife for cutting, our model suffers a performance decrease.
Overall, the results are consistent with previous experiments, demonstrating the effectiveness of our model, trained on a simulation dataset. 
The setup and several evaluation examples are attached in the accompanying video.

\section{Conclusion}
In this work, we investigate the problem named task-oriented 6-DoF grasp pose detection in clutters (TO6DGC). 
This problem considers 6-DoF grasping and cluttered scenarios compared with previous task-oriented grasp pose detection settings. 
To facilitate the development of TO6DGC, we contribute a new dataset, named 6-DoF Task Grasp (6DTG), which features over 2 million 6-DoF task-oriented grasp poses.
We also propose a method, One-Stage Task Grasp (OSTG), and construct several baselines, providing convenience for future research. 
Our OSTG \textit{\textbf{directly}} detects task-oriented grasp poses by adapting a task-oriented point sampling strategy and a task-oriented grasp detection module, outperforming all baselines on multiple metrics. 
Real robot experiments demonstrate the effectiveness of OSTG. 

\section{acknowledgments}
\small{This work was supported partially by NSFC(92470202, U21A20471), National Key Research and Development Program of China (2023YFA1008503), Guangdong NSF Project (No. 2023B1515040025). }


\bibliographystyle{IEEEtran}
\bibliography{main}

\end{document}